\documentclass[10pt,letterpaper]{article}
\usepackage[top=0.85in,left=2.75in,footskip=0.75in,marginparwidth=2in]{geometry}
\usepackage{multirow,tabularx}
\usepackage[utf8]{inputenc}

\usepackage{cite}

\usepackage{nameref,hyperref}

\usepackage[right]{lineno}

\usepackage{microtype}
\usepackage{soul}
\DisableLigatures[f]{encoding = *, family = * }

\raggedright
\usepackage{changepage}

\usepackage[aboveskip=1pt,labelfont=bf,labelsep=period,singlelinecheck=off]{caption}

\makeatletter
\renewcommand{\@biblabel}[1]{\quad#1.}
\makeatother

\usepackage{lastpage,fancyhdr,graphicx}
\usepackage{epstopdf}
\fancyheadoffset[L]{2.25in}
\fancyfootoffset[L]{2.25in}

\usepackage{color}

\definecolor{Gray}{gray}{.25}

\usepackage{graphicx}

\usepackage{sidecap}

\usepackage{wrapfig}
\usepackage[pscoord]{eso-pic}
\usepackage[fulladjust]{marginnote}
\reversemarginpar

\usepackage{ragged2e}

\begin{document}

\vspace*{0.35in}
\justifying
% title goes here:
\begin{flushleft}
{\Large
\textbf\newline{MLT-LE: predicting drug–target binding affinity with multi-task residual neural networks}
}
\newline
% authors go here:
\\
Elizaveta Vinogradova\textsuperscript{1,2},
Karina Pats\textsuperscript{1,2},
Ferdinand Moln\'ar\textsuperscript{2*},
Siamac Fazli\textsuperscript{3*},
\\
\bigskip
\bf{1} Department of Biology, Nazarbayev University, Nur-Sultan, Kazakhstan
\\
\bf{2} Computer Technologies Laboratory, ITMO University, Russia
\\
\bf{3} Department of Computer Science, Nazarbayev University, Nur-Sultan, Kazakhstan
\\
\bigskip
* ferdinand.molnar@nu.edu.kz; siamac.fazli@nu.edu.kz

\end{flushleft}

\section*{Abstract}
Assessing drug-target affinity is a critical step in the drug discovery and development process, but to obtain such data experimentally is both time consuming and expensive. For this reason, computational methods for predicting binding strength are being widely developed. However, these methods typically use a single-task approach for prediction, thus ignoring the additional information that can be extracted from the data and used to drive the learning process. Thereafter in this work, we present a multi-task approach for binding strength prediction. Our results suggest that these prediction can indeed benefit from a multi-task learning approach, by utilizing added information from related tasks and multi-task induced regularization. 

\textbf{Availability of implementation:}
Associated data, pre-trained models, and source code are publicly available at \href{https://github.com/VeaLi/MLT-LE}{https://github.com/VeaLi/MLT-LE}.

% now start line numbers
%\linenumbers

% the * after section prevents numbering
\section*{Introduction}
% One of the early steps in drug discovery is the identification of compounds that have a strong binding to a given target protein often called drug-target affinity (DTA) prediction.
One of the early steps in drug discovery is the identification of compounds that have a strong binding to a given target protein. However, due to high costs and time constraints, it is not possible to screen a large subset of the chemical space by performing laboratory experiments. To solve this problem, many computational methods based on machine learning have been developed~\cite{he_simboost_2017,hu_multi-pli_2021,huang_deeppurpose_2020,nascimento_multiple_2016,nguyen_graphdta_2021,ozturk_deepdta_2018,wang_gandti_2021,wu_moleculenet_2018,abraham}. Most of them are designed to predict one binding constant at a time~\cite{he_simboost_2017,huang_deeppurpose_2020,nguyen_graphdta_2021,ozturk_deepdta_2018}. However, these methods may struggle to adapt to a new drug-target affinity (DTA) prediction pair due to inconsistency and under-representation of training data~\cite{hu_multi-pli_2021,tang_making_2014, liu_algorithm-dependent_2017}. To alleviate this problem, multi-task learning methods have recently been employed~\cite{hu_multi-pli_2021}.

The multi-task approach to molecular property prediction takes advantage of all consistent data by predicting multiple properties simultaneously. By utilizing multiple binding constants at once, multi-task prediction models show improved ability to adapt to new data~\cite{hu_multi-pli_2021,tang_making_2014,ramsundar_massively_2015,zhang_survey_2021,zhang_multi-task_2015,wang_gandti_2021, liu_docking-based_2021}. More technically, multi-task learning is an approach in machine learning that forces a model to favor certain hypotheses over others~\cite{ramsundar_massively_2015,ruder_overview_2017}. Specifically, multi-task learning biases the model toward hypotheses that contribute to the solution of more than one task. Thus, the model's attention becomes focused on the features that really matter, since multiple tasks provide additional evidence for their relevance or irrelevance, thus ideally providing additional regularization. Therefore, in general, models trained to solve multiple tasks are more likely to escape the data-dependent noise and find more useful representations~\cite{ruder_overview_2017}. In addition, models trained to solve multiple related tasks simultaneously show less tendency to overfit, generalize better, and have access to more sample data \cite{zhang_survey_2021,liu_algorithm-dependent_2017,zhang_multi-task_2015, ruder_overview_2017, dahl_multi-task_2014}. 

Multi-task learning can be seen as being inspired by human learning. Humans are able to learn sequentially without forgetting previous knowledge and tend to use the knowledge they already acquired to solve different new tasks~\cite{ruder_overview_2017, kirkpatrick_overcoming_2017}. Initially, multi-task learning was introduced to regularize parameters in linear models and evaluate the task relatedness~\cite{ruder_overview_2017}. Nowadays, multi-task learning is used to find solutions to complex problems involving combinations of multiple data sources, such as in bioinformatics~\cite{mignone_multi-task_2020}. For example, in~\cite{mignone_multi-task_2020}, it was shown that multi-task learning can improve the quality of gene regulatory network reconstruction in settings of scarcely labeled data by using data from multiple organisms simultaneously relying on orthologous genes.

\begin{figure}[ht]
\includegraphics[width=100mm]{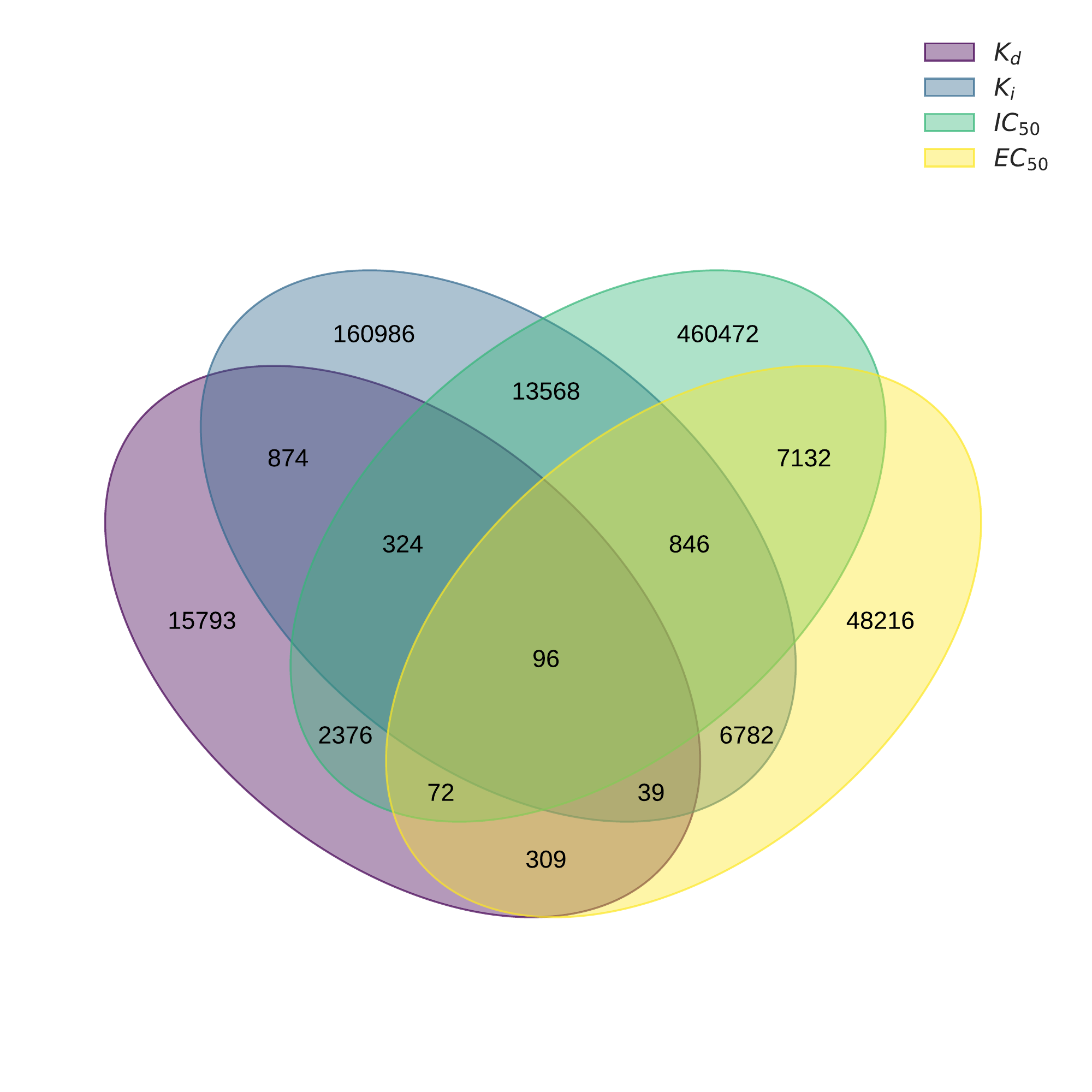}
\captionsetup{labelformat=empty} % makes sure dummy caption is blank
\caption{\color{Gray} \textbf{Figure~\ref{fig0}. The overlap across all four most commonly expressed binding constants for human  drug-target interaction data found in BindingDB is only 96 records (center).}}
\label{fig0}
\end{figure}

The protein-ligand binding strength can be measured in terms of different binding constants. As a result, there exist several independent datasets. Generally, these datasets are used to train different single-task algorithms. Combining this data would naturally produce more powerful prediction models~\cite{ramsundar_massively_2015}. However, only a few multi-task drug target binding strength prediction methods have been developed to date~\cite{hu_multi-pli_2021,wang_gandti_2021}. Why is this? Mainly because only few drug target pairs have several measurements available in terms of binding constants. For example, for a dataset with $\sim717$ thousand records taken from BindingDB, the complete overlap between the four most common binding constants  $\{K_{d} , K_{i}  , IC_{50}  , EC_{50}\}$ is only $96$ records (Fig.~\ref{fig0}).

Despite this, a few multi-task methods for drug-protein binding prediction~\cite{hu_multi-pli_2021,wang_gandti_2021} have shown that they are able to exploit multiple overlaps (intersection) of the data and outperform single-task models in multiple ways~\cite{hu_multi-pli_2021,wang_gandti_2021}.

In this work, we propose a new multi-task framework to address the challenges encountered in these related studies~\cite{hu_multi-pli_2021,wang_gandti_2021}. In particular, we propose an approach that allows using all available data (union), including missing values, and using auxiliary tasks to improve performance. We compare our approach with the state-of-the-art GraphDTA framework, which has been shown to outperform most existing methods~\cite{nguyen_graphdta_2021}.

\section*{Materials and Methods}

\subsection*{Overview of MLT-LE}
\begin{figure}[ht]
\includegraphics[width=\textwidth]{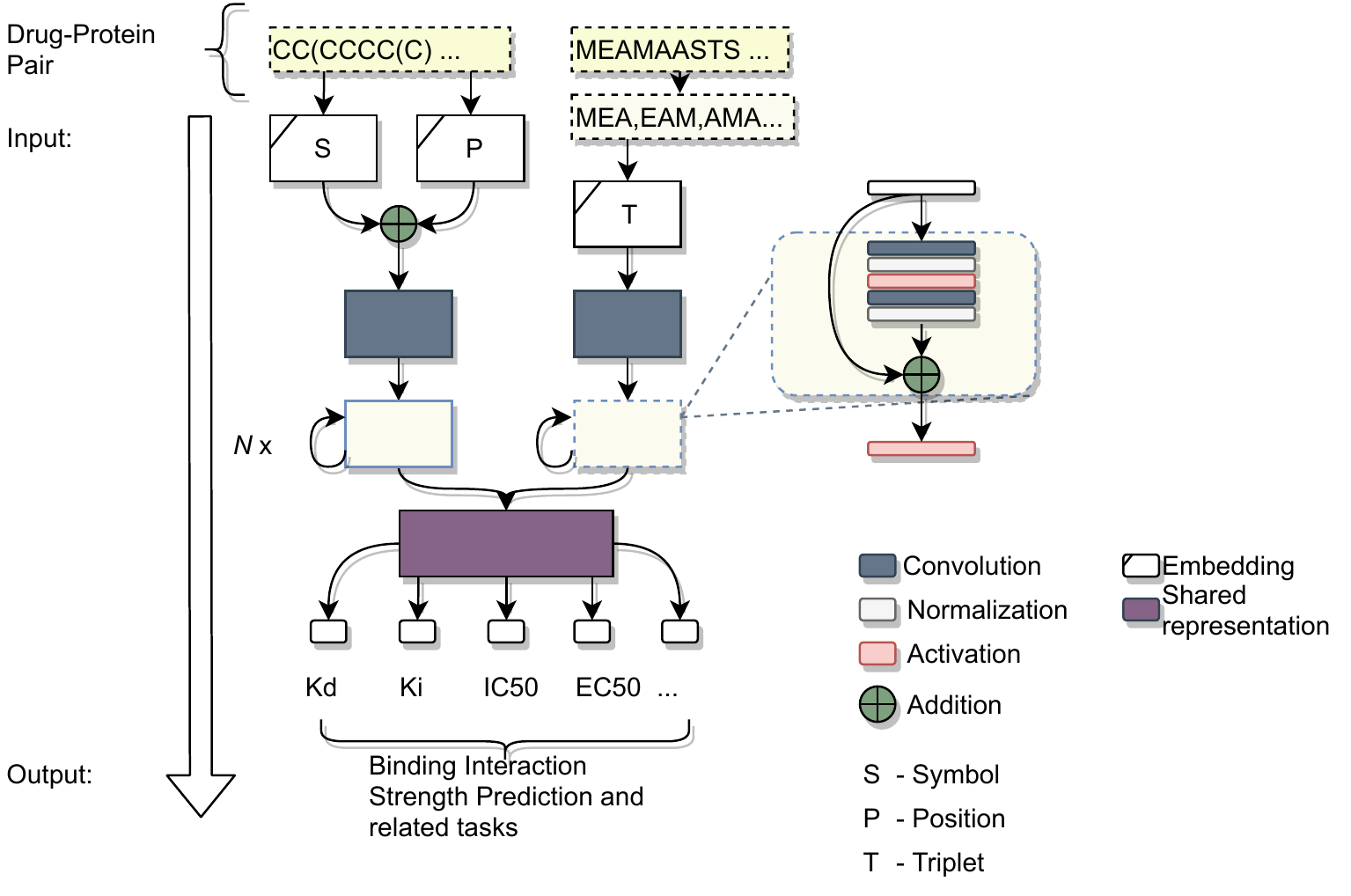}
\caption{\color{Gray} \textbf{General structure of MLT-LE model.} The basic structure of the model within MLT-LE is implemented as a residual convolutional network with adjustable depth and number of outputs.}
\label{fig1} 
\end{figure}

We propose a new deep learning framework - MLT-LE for the prediction of binding strength. Existing methods render the binding strength prediction task as a single-task problem, whereas we propose to solve it in a multi-task manner (Fig. \ref{fig1}). The models in our framework are highly adjustable and support multiple drug and target encodings. 

\subsection*{Handling of missing data}
Combining human data from the BindingDB database~\cite{liu_bindingdb_2007} for any subset of binding constants almost always results in $>50$\% missing data (see table~\ref{tab3} for an example), which makes multi-task approaches challenging to implement. 
%Addressing this situation presents a big challenge. 
Especially since it has to be solved during algorithm training rather than during the preparation phase, since removing the missing data would greatly reduce the sample size. At the same time, filling in the missing data on such a large scale is impossible.
Dahl \emph{et al.} suggested to simply not propagate the error for missing values~\cite{dahl_multi-task_2014} and Hu \emph{et al.} proposed to use branching - propagating the error only for non-missing values using a separate branch in the neural network~\cite{hu_multi-pli_2021}. However, usually, when encountering the problem of training an algorithm with missing data points, a technique called masking is used~\cite{dahl_multi-task_2014}. Masking is the process of dynamically hiding the missing data (i.e. excluding it from the computations) whenever the error is calculated (which is well described for batch optimizers). The error is calculated using some loss function, so masking is part of the loss calculation.
In this work, a different loss adjustment is suggested that shows promising results compared to the masking technique. The developed approach proposes to propagate the last non-missing discounted error found, if there is no data available, thus allowing the use of stochastic optimizers.

\subsection*{Benchmark}
To assess the performance of our models, we compare them to the state-of-the-art GraphDTA framework~\cite{nguyen_graphdta_2021}. We used the model hyperparameters for the GraphDTA models (Table \ref{tab1}), as suggested in the original work~\cite{nguyen_graphdta_2021}, and compared the performance of all models on the same split of training, validation and test data.
Training was conducted for $1000$ epochs and the best models were selected based on the performance on the validation data, particularly by monitoring changes in the mean squared error (MSE) for the target variable(s). Subsequently, all the best performing models were used to obtain predictions for unseen sets of test data.

\begin{table}[!ht]
\centering
\caption{Models hyperparameters.}
\begin{tabular}{lp{0.4\linewidth}ll}
\hline
\textbf{Framework} & \textbf{Optimizer} & \textbf{Learning rate} & \textbf{Batch size} \\ \hline\hline
GraphDTA & Adam & 0.0005 &  1024  \\ \hline
MLT-LE & LookAhead/Nadam with sync. period of 3 & 0.001 & 1024\\ \hline
\end{tabular}
\label{tab1}
\end{table}

\subsection*{Complexity}
Table~\ref{tab2} gives an overview of the complexity of the models involved in the experiments. As can be seen, the total number of parameters for all models is comparable, although some models solve more tasks. The numbers in the model names indicate the number of layers. The plus sign in the model names indicates that the model is trained on merged $K_{d}$+$EC_{50}$ data (see Section "Data"). The prefix 'Res' indicates that the network has residual connections.
For more details on the GraphDTA architecture, please refer to~\cite{nguyen_graphdta_2021}.

\begin{table}[!ht]
\centering
\caption{Model description.}
\begin{tabular}{p{0.12\linewidth}p{0.19\linewidth}p{0.23\linewidth}p{0.13\linewidth}p{0.1\linewidth}p{0.05\linewidth}}
\hline
\textbf{Framework} & \textbf{Model name} & \textbf{Drug representation} & \textbf{Protein representation} & \textbf{Num. of trainable parameters} & \textbf{Num. of tasks} \\ \hline\hline
\multirow{4}{*}{GraphDTA} & GAT2 & Graph attention & Convolution & 1,460,845 & \multirow{4}{*}{ 1 } \\ 
& GATGCN2 & Graph attention - graph convolution combined & & 4,749,573 & \\ 
& GCN3 & Graph convolution  & & 2,062,767 & \\ 
& GIN5 & Graph isomorphism & & 1,297,505 & \\ \hline
\multirow{6}{*}{ MLT-LE } & ResCNN1 & Convolution & Convolution  &  2,391,783 & \multirow{6}{*}{ 7 } \\ 
& ResCNN1+ & & & & \\
& ResCNN1GCN4 & Graph convolution  & &  2,627,895  & \\ 
& ResCNN1GCN4+ & & & & \\
& ResCNN1GIN5 &  Graph isomorphism  & & 2,290,857  & \\ 
& ResCNN1GIN5+ & & & & \\ \hline
\end{tabular}
\label{tab2}
\end{table}

\subsection*{Data} \label{sec:Data}
In this work, we used human $K_{d}$ and $EC_{50}$ data from BindingDB v2022m3~\cite{liu_bindingdb_2007}, each divided into $3$ sets of the same size, namely train, validation, and test, to mitigate the effects of overfitting.
The original subset of human data from BindingDB contained $1,274,797$ records, with $525,152$ unique SMILES and $1,981$ unique amino acid sequences of target proteins. Only records with at least one binding constant were kept. Then, the records were filtered to contain only binding constants recorded without range boundaries, indicated by $+ -$ and $><$ signs. The SMILES sequences were then canonicalized using RDkit-pypi v2022.3.5~\cite{noauthor_rdkit_nodate}, and the isomeric information was removed. Only records with valid SMILES strings were kept. As a result, the dataset contained slightly fewer unique SMILES sequences, $498,742$. The QED coefficient~\cite{bickerton_quantifying_2012} was then calculated for each SMILES using RDkit (to be used as an auxiliary task later). The individual data records were then aggregated using the median, as was done in~\cite{tang_making_2014}. Only valid protein sequences starting with a start codon were considered. Then, the individual records were labeled as ``active'' or ``non-active'' drug-target pairs using a threshold of $(K_{d} | K_{i}  | IC_{50}  | EC_{50}) <1\mu M$ for binarization, as done in~\cite{moret_generative_2020, olivecrona_molecular_2017} or similarly in~\cite{grisoni_combining_2021}. After that, the training dataset contained a total of $716,342$ records. The overlap of records containing different binding constants was then calculated (Fig. \ref{fig0}); the total overlap between records was $96$ records in total, with missing values corresponding to more than $>77\%$. Next, only records with $K_{d}$ or $EC_{50}$ non-missing values were kept, yielding $19,796$ $K_{d}$ records and  $63,487$ $EC_{50}$ records. The data were then shuffled and divided into three equal parts for each set, resulting in $6,598$ training, $6,598$ validation, and $6,600$ test records for $K_{d}$ and $21,153$, $21,153$, $21,155$ records for $EC_{50}$ respectively, all of which contain non-overlapping drug-target pairs. In addition, we created a combined $K_{d}$+$EC_{50}$ dataset, denoted in this work by ''+'', by combining the training and validation datasets from the separate $K_{d}$ and $EC_{50}$ data. It should be noted that we also saved the target pH information for all datasets in order to use it as another auxiliary task.
Table~\ref{tab3} shows the number of non-missing values in the train data for each dataset. Note that the validation and test datasets have a similar distribution of records as the corresponding train dataset.

\begin{table}[!ht]
\centering
\caption{Number of non-missing values in train data.}
\begin{tabular}{p{0.18\linewidth}p{0.08\linewidth}p{0.08\linewidth} p{0.08\linewidth}p{0.08\linewidth}p{0.08\linewidth}p{0.08\linewidth}p{0.08\linewidth}}\hline
\textbf{Dataset} & $\mathbf{K_{d}}$ & $\mathbf{K_{i}}$ & $\mathbf{IC_{50}}$ & $\mathbf{EC_{50}}$ & \textbf{active} & \textbf{pH} & \textbf{QED }\\ 
\hline
\hline
$K_{d}$ set & 6,598 & 11 & 20 & 1 & 6,598 & 580 & 6,598 \\
$EC_{50}$ set & 3 & 159 & 201 & 21,153 & 21,153 & 1,144 & 21,153 \\
Combined ('$+$') $K_{d}+EC_{50}$ set & 6,600 & 170 & 221 & 21,153 & 27,750 & 1,723 & 27,750 \\ \hline
\end{tabular}
\label{tab3}
\end{table}

\section*{Results}

\begin{figure}[ht]
\begin{adjustwidth}{-1in}{0in}
\begin{flushright}
\includegraphics[width=163mm]{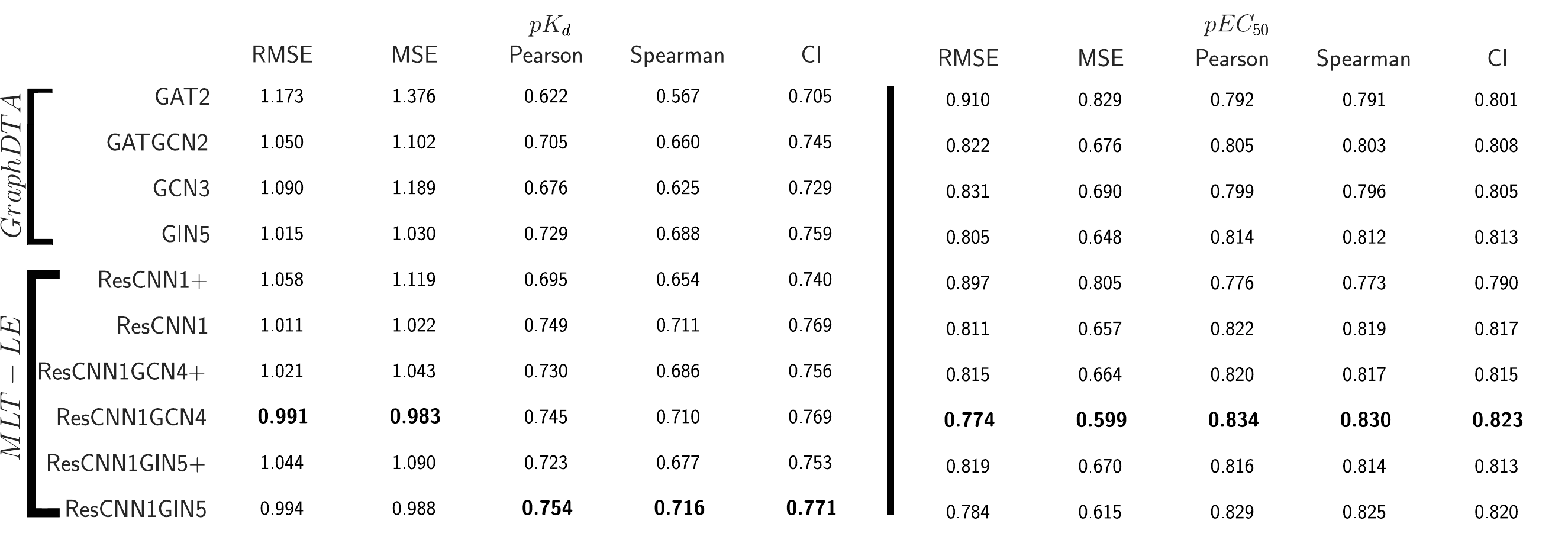}
\end{flushright}
\caption{\color{Gray} \textbf{Comparison of the model performances on the test set for $K_{d}$ and $EC_{50}$ data}. The best scores obtained for each metric are shown in bold.}
\label{fig2}
\end{adjustwidth}
\end{figure}

Figure~\ref{fig2} shows the performance of the best models obtained on the test set. It can be seen that the MLT-LE models showed comparable or better performance than the baseline GraphDTA models on all metrics. Additional statistical testing of the observed difference between the predictions of the different frameworks can be found in the supplementary materials. The best scores obtained for each metric are shown in bold.

\begin{figure}[ht]
\includegraphics[width=\textwidth]{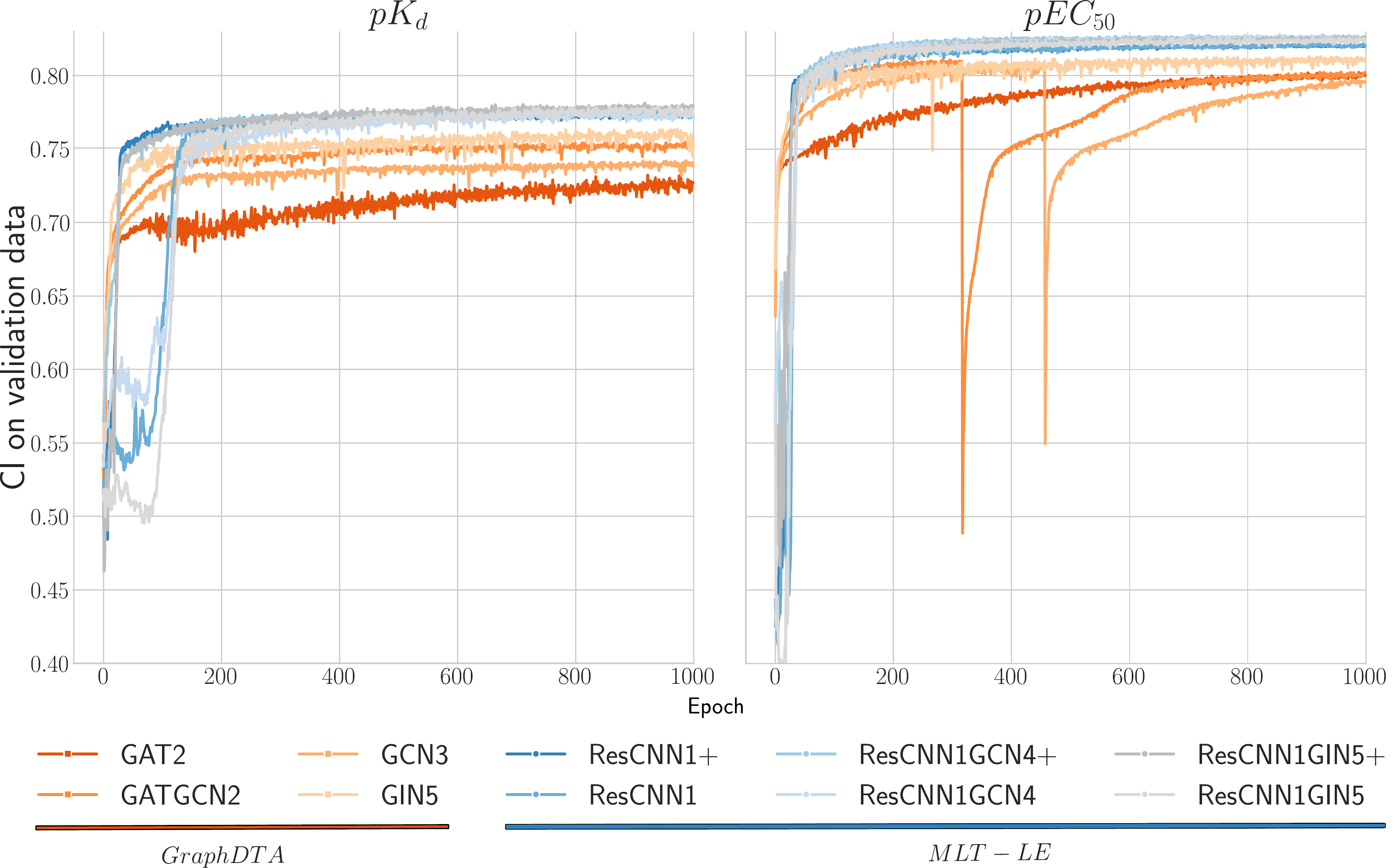}
\caption{\color{Gray} \textbf{Comparison of model performance on the validation data for $K_{d}$ and $EC_{50}$ data}. GraphDTA performance is shown in orange and MLT-LE performance is shown in blue.}
\label{fig3}
\end{figure}

Figure~\ref{fig3} shows the performance of all models used in the experiment on the validation data. It can be seen that the GraphDTA and MLT-LE framework models show comparable performance, although the GraphDTA models show more stable convergence, while MLT-LE achieves both the highest concordance index ($CI$) and solves more tasks. 

It was also observed that the performance of MLT-LE models is also of high quality for other tasks; more information on this is provided in the supplementary material.

In addition, similar model performance was observed when training with a smaller batch size of $256$ over $100$ epochs for the same datasets (the results of this experiment are shown in the supplementary material). 

%\clearpage makes sure that all above content is printed at this point and does not invade into the upcoming content
%\clearpage

\section*{Discussion}
Incorporating data from various sources helps to obtain more accurate predictions of molecular properties~\cite{ramsundar_massively_2015}. However for this task, the combination of data sources results in a substantial amount of missing data (as can be seen in Table \ref{tab3}). Thus, the question may arise whether combining these data would be beneficial for the prediction~\cite{wang_gandti_2021,liu_docking-based_2021}. The results of the present work suggest that missing data can be included and is shown to be beneficial in terms of performance (Fig.~\ref{fig2}). We attribute this observation to the effect of multi-task models being able to incorporate data from multiple sources and thus utilizing information that is not available for single-task models.

\section*{Conclusion}
In this work, we compared our binding strength prediction framework - MLT-LE with the state-of-the-art GraphDTA framework. We tested the performance on two different subsets of BindingDB. We found that MLT-LE performs well for the benchmark datasets across all metrics. We furthermore observed that a multi-task convolutional model can provide comparable results to a single-task graph model, and that a multi-task graph model can perform better than a single-task graph model. We also observed the beneficial effects of incorporating more data, even given the high percentage of missing values.

We believe that it may be possible to further improve performance on the binding strength prediction task using data from more organisms if the orthologous protein sequences in these organisms are matched properly and if a special task is used in the neural network to distinguish the organisms, which can smooth out the effects of different medians of binding constants in these organisms.

\section*{Supporting Information}
Associated and supplementary data, data preparation routines, pre-trained models, and source code are publicly available at \href{https://github.com/VeaLi/MLT-LE}{https://github.com/VeaLi/MLT-LE}. 

\section*{Acknowledgments}
The co-authors would like to acknowledge the support of ITMO University Research Grants funding to EV and KP (621314) and Nazarbayev University Research Grants funding to SF and FM (240919FD3926, 091019CRP2108 and 110119FD4520).

\nolinenumbers

%This is where your bibliography is generated. Make sure that your .bib file is actually called library.bib
\bibliography{ms}
\bibliographystyle{IEEEtran}

\end{document}

\bibliographystyle{IEEEtran}

\vspace*{0.35in}
\justifying
% title goes here:
\begin{flushleft}
{\Large
\textbf\newline{Supplementary Material for ``MLT-LE: predicting drug–target binding affinity with multi-task residual neural networks''}
}
\newline
% authors go here:
\\
Elizaveta Vinogradova\textsuperscript{1,2},
Karina Pats\textsuperscript{1,2},
Ferdinand Moln\'ar\textsuperscript{2*},
Siamac Fazli\textsuperscript{3*},
\\
\bigskip
\bf{1} Department of Biology, Nazarbayev University, Nur-Sultan, Kazakhstan
\\
\bf{2} Computer Technologies Laboratory, ITMO University, Russia
\\
\bf{3} Department of Computer Science, Nazarbayev University, Nur-Sultan, Kazakhstan
\\
\bigskip
* ferdinand.molnar@nu.edu.kz; siamac.fazli@nu.edu.kz

\end{flushleft}

Figures \ref{fig1}-\ref{fig2} below show a comparison of model performance on validation and test data when training for 100 epochs with a smaller batch size of 256. This experiment uses the same data partitions and predictive models as the main study. As can be seen in Figures \ref{fig1}-\ref{fig2} the performance of the models can be considered consistent for the two frameworks for two different batch sizes.

\begin{figure}[ht]
\includegraphics[width=\textwidth]{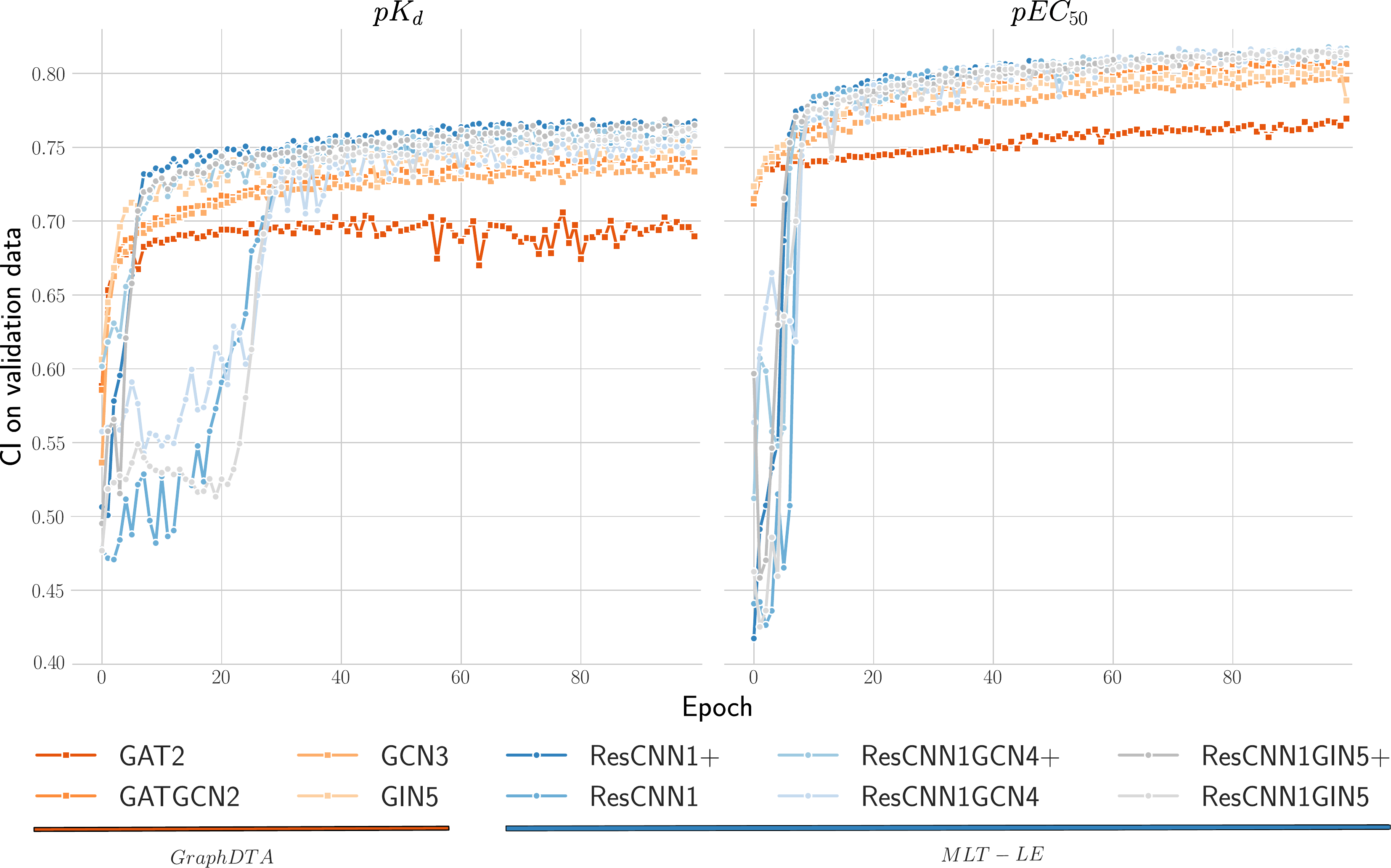}
\caption{\color{Gray} \textbf{Comparison of model performance on the validation data for $K_{d}$ and $EC_{50}$ data when training in 100 epochs with a smaller batch size of 256}. GraphDTA performance is shown in orange and MLT-LE performance is shown in blue.}
\label{fig1}
\end{figure}

\begin{figure}[ht]
\begin{adjustwidth}{-1in}{0in}
\begin{flushright}
\includegraphics[width=163mm]{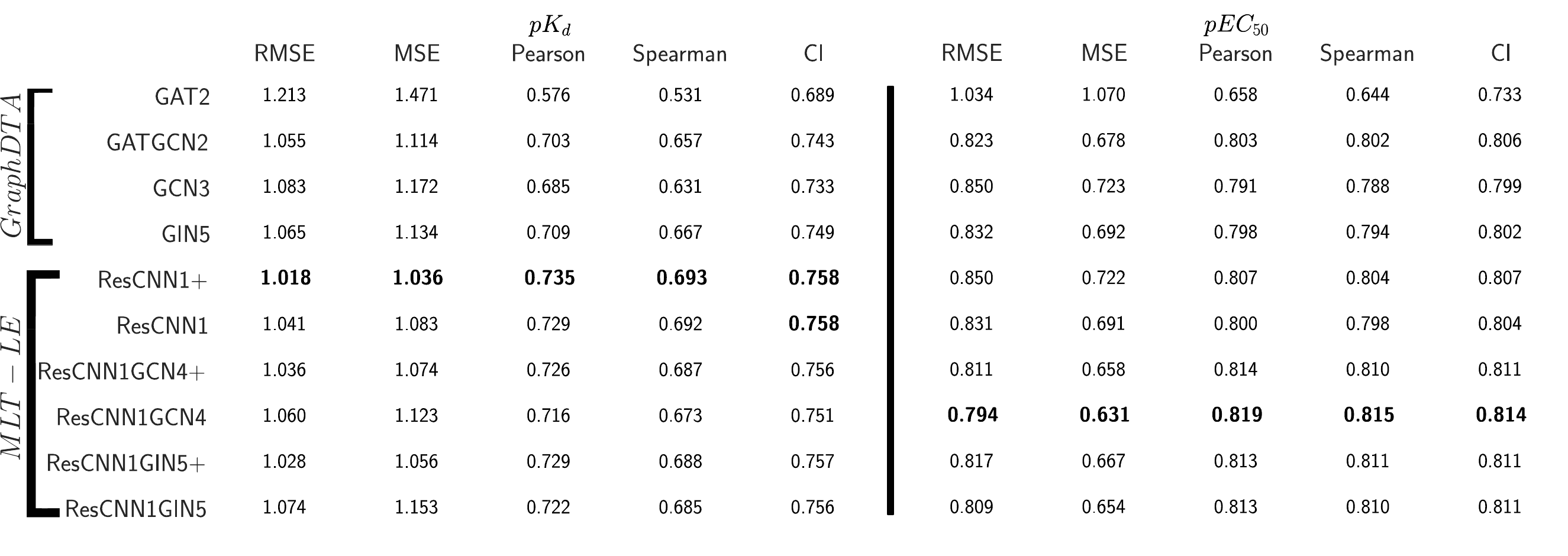}
\end{flushright}
\caption{\color{Gray} \textbf{Comparison of the performances 
of the best models on the test set for $K_{d}$ and $EC_{50}$ data when training in 100 epochs with a smaller batch size of 256}. The best scores obtained for each metric are shown in bold.}
\label{fig2}
\end{adjustwidth}
\end{figure}

\begin{figure}[ht]
\includegraphics[width=\textwidth]{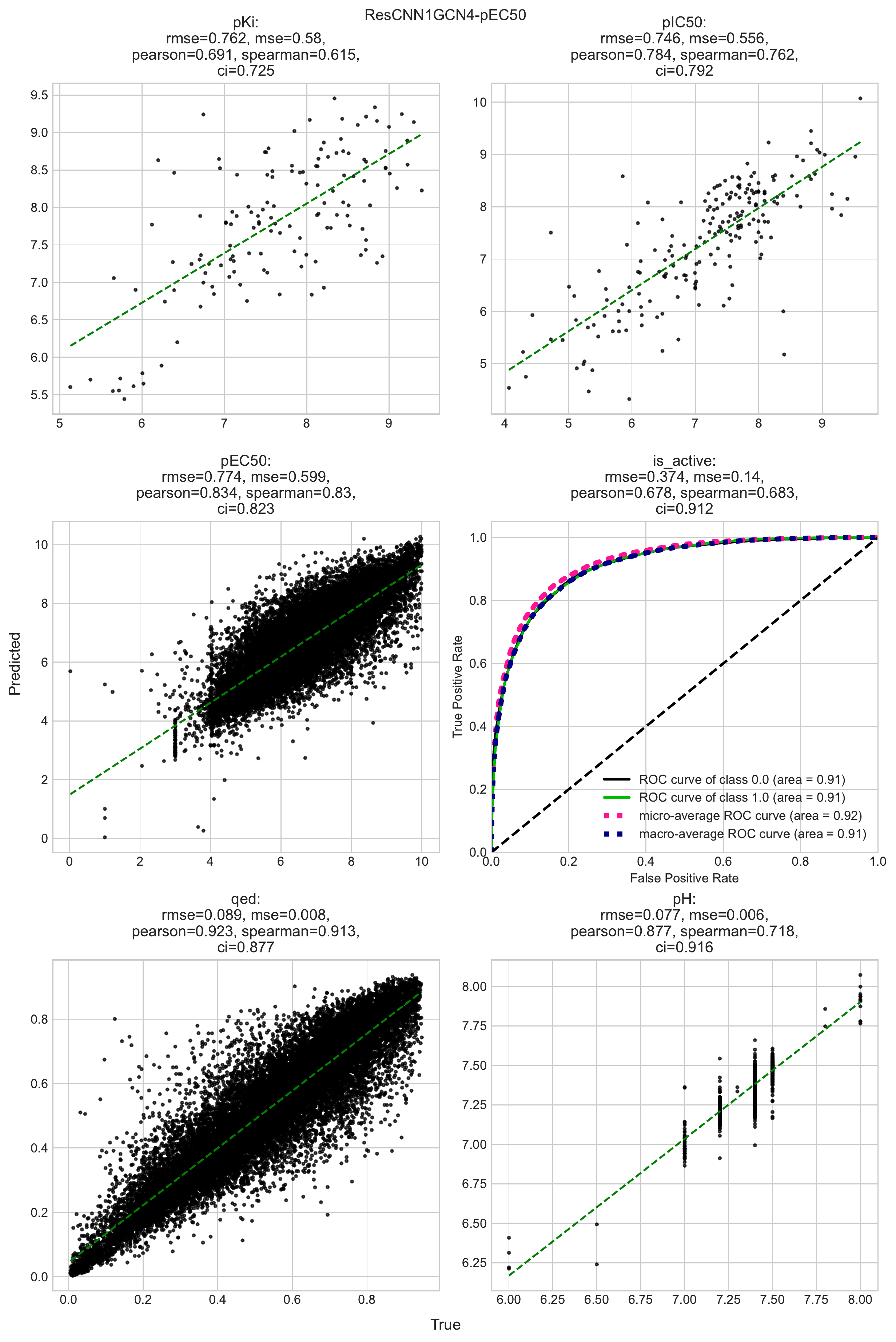}
\caption{\color{Gray} \textbf{Performance of ResCNN1GCN4 model $EC_{50}$ subset test data }}
\label{fig3}
\end{figure}

Figure \ref{fig3} shows the performance of the best MLT-LE model (ResCNN1GCN4) for all tasks $EC_{50}$ subset test data.

\subsection*{Statistical significance test of performance on $EC_{50}$ data}

\begin{figure}[ht]
\includegraphics[width=100mm]{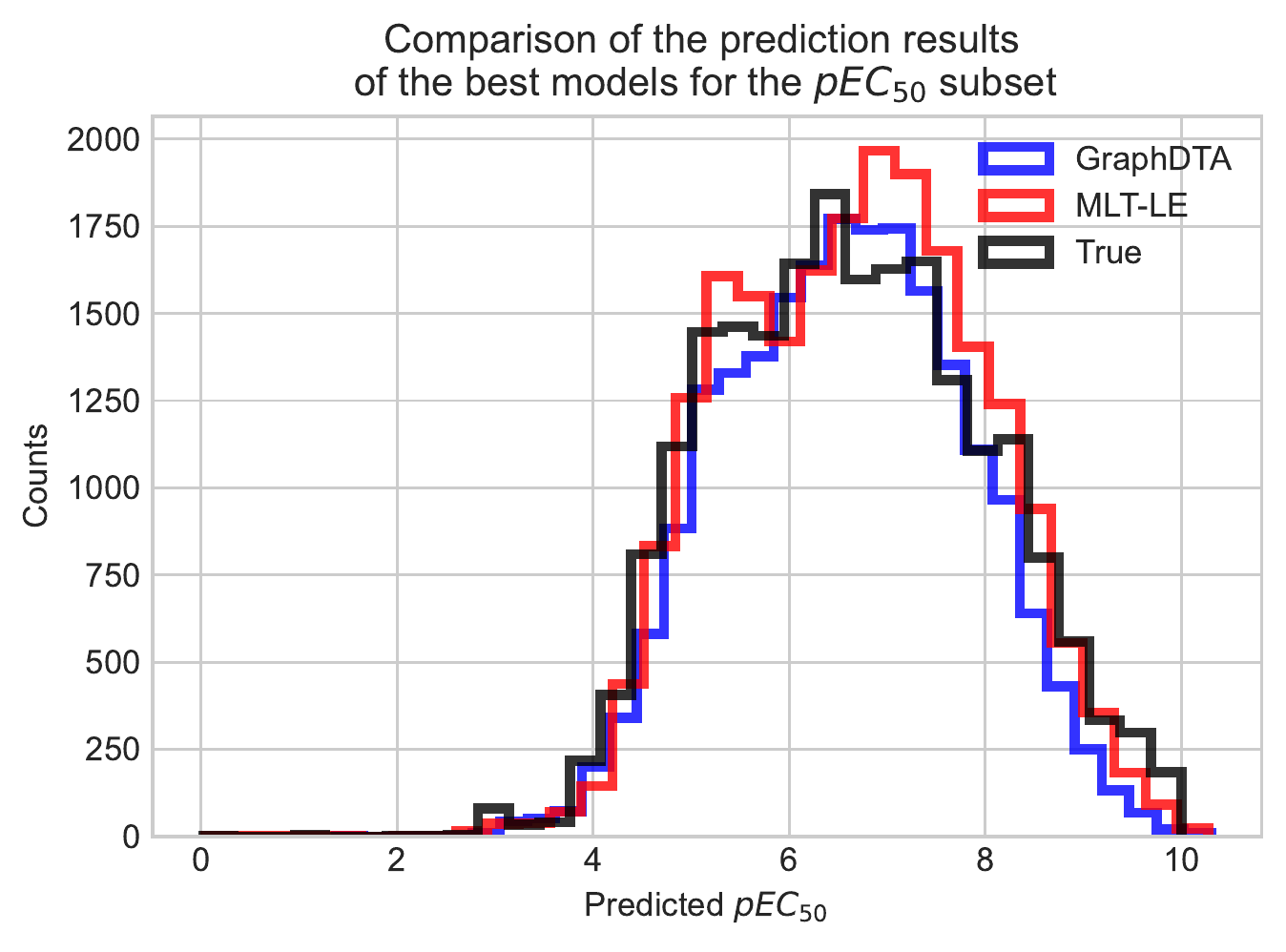}
\caption{\color{Gray} \textbf{Distribution of predicted values compared to actual distribution}}
\label{fig4}
\end{figure}

It was observed that the prediction accuracy of the GraphDTA models is lower than the prediction accuracy of the MLT-LE models in the test sets. The distribution of predictions between the two best models from each framework for the $EC_{50}$ dataset can be seen in Fig.\ref{fig4}.

In Table \ref{tab_given}, we verify that the observed difference in predicted values between the models is statistically significant using a sign test.

\begin{table}[!ht]
\centering
\caption{Significance testing}
\begin{tabular}{p{0.4\linewidth}|p{0.5\linewidth}} \hline
\multicolumn{2}{l}{\textbf{Given:}} \\  \hline
N = 21154 & Number of predicted values \\ \hline
\multicolumn{2}{l}{\textbf{Two-sided hypothesis:}} \\ \hline
\multicolumn{2}{l}{\textbf{H$_0$:} There is no difference between GraphDTA and MLT-LE values} \\
\multicolumn{2}{l}{\textbf{H$_1$:} There is difference} \\ \hline
\multicolumn{2}{l}{\textbf{Number of positive and negative differences for sign-test:}} \\ \hline
11999 - negative differences & Difference between matched pairs of predictions \\
9155 - positive differences  \\ \hline
\multicolumn{2}{l}{\textbf{Binomial test on pairs of differences:}} \\ \hline
Number of trials = 21154\\
Number of successes = 11999\\
Probability of success = 50\%\\
P-value: 0.25E-84 & \\ \hline
\multicolumn{2}{l}{\textbf{Conclusion}} \\ \hline
\multicolumn{2}{l}{The null hypothesis can be rejected with a p-value of 0.25E-84.}\\
\multicolumn{2}{l}{The results suggest that the observation of such a difference between}\\
\multicolumn{2}{l}{pairs of observations or more extreme is unlikely.}\\ \hline
\end{tabular}
\label{tab_given}
\end{table}

The results of the performed test indicate that the observed difference in performance between the two frameworks in our experiment may be considered significant.